\documentclass[sigconf,authorversion,nonacm]{acmart}

\AtBeginDocument{%
  }

\usepackage{graphicx}

\usepackage{amsfonts}
\usepackage{amsmath}
\usepackage{hyperref}

\usepackage{enumitem}
\setlist{leftmargin=4.0mm}

\usepackage{algorithm}
\usepackage{algorithmic}
\usepackage{balance}

\begin{document}

\title{Simulating the Economic Impact of Rationality through Reinforcement Learning and Agent-Based Modelling}


\author{Simone Brusatin}
\affiliation{%
  \institution{Università di Trieste}
    \country{Italy}
}
\orcid{0009-0001-0078-7208}

\author{Tommaso Padoan}
\affiliation{%
  \institution{Università di Trieste}
  \country{Italy}
}
\orcid{0000-0001-7814-1485}

\author{Andrea Coletta}
\affiliation{%
  \institution{Banca d'Italia$^{\dag}$}
    \country{Italy}
}
\orcid{0000-0003-1401-1715}

\author{Domenico Delli Gatti}
\affiliation{%
  \institution{Università Cattolica del Sacro Cuore}
    \country{Italy}
}
\orcid{0000-0001-8819-090X} 

\author{Aldo Glielmo}
\affiliation{%
  \institution{Banca d'Italia$^{\dag}$}
    \country{Italy}
}
\authornote{Corresponding author, aldo.glielmo@bancaditalia.it. \\
 This article was published in the \emph{Proceedings of the Fifth ACM International Conference on AI in Finance}, and is available also at \url{https://doi.org/10.1145/3677052.3698621} \\
 $^\dag$The views and opinions expressed in this paper are those of the authors and do not necessarily reflect the official policy or position of Banca d’Italia.
\\
\\
\\
\\
 }
\orcid{0000-0002-4737-2878}

\renewcommand{\shortauthors}{Brusatin, Padoan, Coletta, Delli Gatti, Glielmo}

\begin{abstract}
Agent-based models (ABMs) are simulation models used in economics to overcome some of the limitations of traditional frameworks based on general equilibrium assumptions.
However, agents within an ABM follow predetermined `bounded rational' behavioural rules which can be cumbersome to design and difficult to justify.
Here we leverage multi-agent reinforcement learning (RL) to expand the capabilities of ABMs with the introduction of `fully rational' agents that learn their policy by interacting with the environment and maximising a reward function.
Specifically, we propose a `Rational macro ABM’ (R-MABM) framework by extending a paradigmatic macro ABM from the economic literature.
We show that gradually substituting ABM firms in the model with RL agents, trained to maximise profits, allows for studying the impact of rationality on the economy.
We find that RL agents spontaneously learn three distinct strategies for maximising profits, with the optimal strategy depending on the level of market competition and rationality.
We also find that RL agents with independent policies, and without the ability to communicate with each other, spontaneously learn to segregate into different strategic groups, thus increasing market power and overall profits.
Finally, we find that a higher number of rational (RL) agents in the economy always improves the macroeconomic environment as measured by total output.
Depending on the specific rational policy, this can come at the cost of higher instability.
Our R-MABM framework allows for stable multi-agent learning, is available in open source, and represents a principled and robust direction to extend economic simulators.
\end{abstract}





\maketitle

\begin{figure*}[h!]
  \centering
\includegraphics[width=1.02\textwidth]{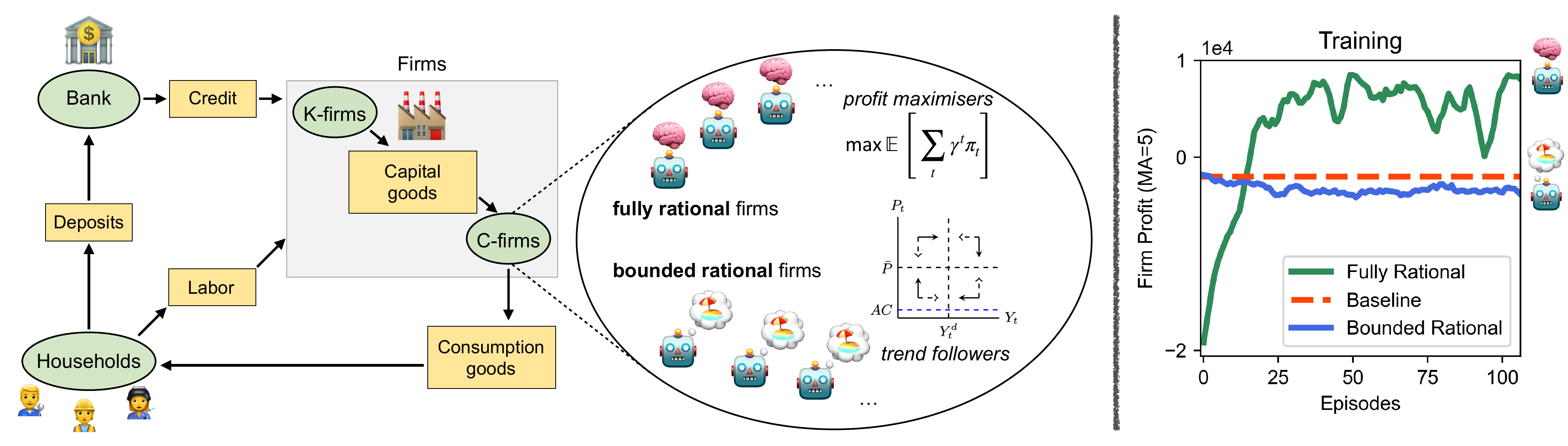}
   \vspace{-0.15in}
      \caption{\textbf{The R-MABM model.} 
  \normalfont
  The left panel shows a schematic diagram of the basis model, in which 4 types of agents (green ovals) exchange goods (yellow rectangles). The arrows represent the flow of the specific good, from provider to receiver.
  The middle panel illustrates the extension of the basis model we implement with our R-MABM framework.
  Consumption-good producing firms `C-firms' in the standard model are exclusively `bounded rational' and take decisions using a heuristic trend-following rule. 
  These are augmented with `fully rational' RL agents that take decisions in order to maximise profits.
  The right panel shows a typical learning curve where fully rational RL agents learn to accumulate higher profits than bounded rational agents as the number of learning episodes progresses.
  }
  \vspace{-0.1cm}
  \label{fig:training_reward}
\end{figure*} 

\vspace{-0.1cm}
\section{Introduction}
Traditional economic simulation models are mostly based on general equilibrium frameworks, and are far from being a perfect representation of real economies, due to their strong assumptions, which typically do not allow for agent heterogeneity, bounded rationality or nonequilibrium dynamics~\cite{farmer2009economy,dawid2018agent}.
Today, advanced computational methods hold the promise of solving some of these key limitations~\cite{dawid2008computational,rothe2015economics,blume2015introduction,benedetti2022black,monti2023learning}. 
Specifically, multi-agent systems (MAS)~\cite{wooldridge2009introduction} may offer a bottom-up approach to model economic systems that naturally addresses the above mentioned limitations of traditional models. 
Since the early 2010s, MAS have gradually been adopted and gained popularity in economics under the name of ``agent-based models'' (ABMs)~\cite{farmer2009economy,dawid2018agent,axtell2022agent}. 
ABMs can model economic systems by simulating a number of individual economic agents, representing decision makers and institutions, which can be defined to be heterogeneous and act with bounded rationality in non-equilibrium environments~\cite{tesfatsion2023agent,tesfatsion2006agent}.
%
These agents typically rely on a pool of hand-crafted behavioural rules, that are defined in advance by economists to mimic the complexity of real economies~\cite{vyetrenko2020get}. 
While this approach allows addressing some of the limitations described, it also poses new challenges, such as the difficulty in defining realistic behaviours, and the absence of learning agents and of `rational' agents that optimise their behaviour based on a specific objective.

Reinforcement Learning (RL)~\cite{sutton2018reinforcement} can offer a solution to classical ABM limitations as, in principle, it can allow for a grounded introduction of rational optimisation and learning in simulation models, without the need for the over-simplistic assumptions of traditional economic modelling frameworks.
Moreover, RL can facilitate the design of accurate economic ABMs, as it can eliminate the burden of programming detailed behavioural rules in favour of the easier task of choosing meaningful reward functions, typically known as `utility functions' in the economic literature.

\vspace{0.1cm}
\textbf{Our contribution}. In this work, we contribute towards this interdisciplinary challenge by proposing a framework to leverage multi-agent RL to expand the capabilities of classical macro ABMs.
Specifically, we extend and study the well-known macroeconomic ABM with capital and credit (`CC-MABM') from Assenza et al.~\cite{assenza2015emergent}. 
Such a model considers an economy populated by households, firms and banks. 
Households supply labour to firms, consume goods, and save money through deposits at the banks, and banks provide loans to firms. 
Our extended model substitutes a variable number of firms with RL agents. 
Importantly, we refer to the firms of the original models as `bounded rational' in the sense that they behave according to fixed rules that are not the result of a maximisation of a utility function, while we refer to RL firms as `fully rational' since they act in order to explicitly maximise profits.
From these considerations, we name our framework \textit{`Rational macro ABM’} (R-MAMB).
Figure~\ref{fig:training_reward} shows a visualisation of the R-MAMB model, and a sample of the learning process of rational firms.

We experimentally evaluate our framework with extensive simulations, and show that RL agents spontaneously learn three distinct strategies for maximising profits, with the optimal strategy depending on the level of market competition and rationality in the economy.
We also find that agents with \textit{independent} policies spontaneously learn to segregate into different strategic groups, thus increasing market power and overall profits, without explicitly communicating with each other. 
Finally, we assess the macroeconomic impact of higher degrees of rationality by measuring the direct effects on GDP and on its volatility.
To the best of our knowledge, R-MABM is the first attempt at extending macroeconomic ABMs with the introduction of rational agents using RL.
Our results amply demonstrate the robustness of the R-MABM framework and its relevance to a vast community at the intersection between economics and computer science, thus paving the way for numerous applications and future investigations. 
Accordingly, we made the code for the R-MABM easy to use and freely available \footnote{The R-MABM is available in open source at \url{https://github.com/Brusa99/R-MABM}, while the code used for the basis model is available at \url{https://github.com/bancaditalia/ABCredit.jl}.}.

\vspace{0.1cm}
\textbf{Related work.} 
Thanks to recent breakthroughs in deep-RL and multi-agent RL (MARL)~\cite{gronauer2022multi,albrecht2023multi}, along with the ever-growing availability of computational power, RL and MARL are rapidly emerging as attractive powerful tools to expand the capabilities of traditional ABMs and pave the way to a new generation of economic simulators~\cite{charpentier2021reinforcement,ardon2023phantom,dwarakanath2024abides}.
Several works have investigated the use of RL towards economic modelling for different environments and settings~\cite{atashbar2022deep}, including learning monetary policies ~\cite{hinterlang2021optimal}, calibrating economic ABMs~\cite{glielmo2023reinforcement},
solving heterogeneous general equilibrium models~\cite{hill2021solving},
approaching consumption-saving problems~\cite{kuriksha2021economy}, studying social segregation dynamics~\cite{sert2020segregation} or mortgage relief strategies  \cite{garg2024heterogeneous},
modelling agents with heterogeneous bounded rationality~\cite{evans2024learning}, or modelling decision making process in technology uptake models~\cite{klugl2023modelling}.
Another very active line of research focuses on the use of RL in the simulation of financial stock markets ~\cite{vadori2024towards,coletta2022learning,albers2024beliefs}.

For models of the entire economy, a recent example is the work of Johanson et al.~\cite{johanson2022emergent}, which shows that basic economic phenomena of microeconomics can emerge spontaneously from learning agents.
Other work focuses on real business cycle (RBC) models showing that optimal policies can be learnt through MARL~\cite{zheng2022ai,curry2023learning,mi2023taxai,lindau2023optimal,atashbar2023ai}, using different learning techniques including \textit{curriculum learning}~\cite{bengio2009curriculum}. 
In \cite{mi2023taxai}, Mi et al. showcase the advantages of MARL systems over traditional models in designing and studying taxation policies. 

As we focus on RL modelling of firms in our macro ABM for this study, our work is also related to the vast literature on algorithmic pricing using RL, pioneered by Tesauro and Kephart~\cite{tesauro1998foresight,tesauro2002pricing} and more recently taken up by many studies in both economics and computer science
\cite{waltman2008q,cai2018reinforcement,calvano2020artificial,kastius2022dynamic}.

Finally, our work is related to \cite{dosi2020rational}, where Dosi et al. study the effects of substituting the traditional bounded rational behavioural rules of a macroeconomic ABM with more complex adaptive rules, without using RL.

\vspace{-0.1cm}
\section{The Rational-MABM model} 

Our work expands a traditional macroeconomic ABM introducing fully rational agents through reinforcement learning. 
This section first provides an overview of the MABM that forms the basis of our framework, which we also call the `basis model', and then details the multi-agent RL scheme we develop to extend it.

\subsection{The basis model}
We here briefly describe the basis macro ABM, namely the macroeconomic ABM with capital and credit (CC-MABM) from Assenza et al~\cite{assenza2015emergent}.
For a more detailed description of the model and its parameters we refer to the original manuscript.

\vspace{0.1cm}
\textbf{Agent types.}
The basis model comprises three types of agents defined as follows.
\begin{itemize}
    \item \textbf{Households} consist of workers and capitalists. 
    Workers supply labour, receiving a wage.
    Capitalists receive dividends and do not work.
    Both deposit accumulated wealth in banks.
    \item \textbf{Banks} receive household deposits and extend loans to firms.
    \item \textbf{Firms} consist of K-firms and C-firms.
    K-firms require labour for their production, and produce capital goods i.e., machinery and equipment, that they sell to C-firms.
    C-firms require both labour and capital goods for their production and produce consumption goods that they sell to households.
\end{itemize}
Figure \ref{fig:training_reward} summarises the agents and their interactions. 
For the sake of brevity, in the following we provide extra details only on the specific behavioural rules of households and C-firms that are most relevant to our later experiments.

\vspace{0.1cm}
\textbf{Markets.}
All of the households aim to buy consumption goods, and therefore participate in a \textit{search and matching} consumption market, defined as follows.
\begin{itemize}
    \item At the start of period $t$, each consumer determines their consumption budget, based on their income and bank deposits.
    \item To decide where to buy the goods, each consumer visits $z_c$ randomly selected C-firms and sorts them by their retail price from lowest to highest.
    \item The consumer starts buying goods from the first firm and, if the consumption budget is not exhausted, the consumer moves on to the second firm in the order, and so on.
\end{itemize}
Note that, a higher (lower) value of $z_c$ implies that each consumer will be able to compare the prices of a higher (lower) number of firms.
Hence, importantly, we can consider those this parameter as an effective controller of the level of competition of the search and matching market described.

\vspace{0.1cm}
\textbf{Price and quantity decisions of firms.} 
The model assumes that at each time step $t$, a firm $i$ decides a price $P_{i, t}$ at which to sell its goods, and a target quantity $Y^*_{i,t}$ of goods it aims to produce. 
Based on the desired production $Y^*_{i,t}$, the firm decides how many workers to employ $N_{i,t}$ and how much capital $K_{i,t}$ is required.
The firm will then attempt to acquire labour and capital in the corresponding markets.
Finally, the production function follows a Leontief technology, i.e., $Y_{i,t} = \min\ (\alpha_{N} N_{i,t}, \alpha_{K} K_{i,t})$ where $\alpha_{N}$ and $\alpha_{K}$ are labour and capital productivity, respectively and $Y_{i,t}$ is the quantity of goods actually produced.
Notice that consumption goods are non-storable, so unsold goods do not carry over to the next period.

For the rest of this work, we will often call a defined policy to set $P_{i, t}$ and $Y^*_{i,t}$ a `price-quantity strategy', or simply a `strategy'.
The price-quantity strategy of C-firms in the basis model is based on the following.
At the end of the period $t$, the firm $i$ observes the amount of goods sold $Y^s_{i, t} = \min \left(Y_{i, t}, Y_{i, t}^d\right)$, where $Y_{i, t}^d$ is the actual demand, and the average price $P_t$ charged by competitors. 

To set future prices and quantities, firms use the past step \emph{firm stock} $\Delta_{i, t}^Y = Y_{i, t} -  Y_{i, t}^d$ and \emph{price delta} $\Delta_{i, t}^P = P_{i, t} -  P_{t}$.
The firm stock is the difference between the amount of goods produced, and the amount of goods demanded.
If positive, it amounts to the unsold `stock' of goods piled up by the firm in a given step.
If negative, its absolute value equals the quantity of extra goods that could have been produced to meet the demand.
The price delta is the distance between the firm's price and the average market price, positive if above the average or negative if below it.
C-firms in the basis model adjust prices and target quantities following the simple heuristic pictured in Figure \ref{fig:training_reward} and described by the following equations
\begin{align}
     P_{i,t+1} &= \left\{\!\!
    \begin{array}{ll}
        P_{i,t}(1+\eta_{i,t+1}) & \text{ if }\Delta_{i,t}^Y \le 0 \land \Delta_{i, t}^P < 0 \\[2mm]
        P_{i,t}(1-\eta_{i,t+1}) & \text{ if }\Delta_{i,t}^Y > 0 \land \Delta_{i, t}^P \ge 0 \\[2mm]
        P_{i,t} & \!\text{ otherwise}
    \end{array}\right.
\label{eq:adg-quantity-setting} \\
    Y^*_{i,t+1} &= \left\{\!\!
    \begin{array}{ll}
        Y_{i,t} + \rho |\Delta_{i,t}^Y| & \text{ if }\Delta_{i,t}^Y \le 0 \land \Delta_{i, t}^P \ge 0 \\[2mm]
        Y_{i,t} - \rho |\Delta_{i,t}^Y| & \text{ if }\Delta_{i,t}^Y > 0 \land \Delta_{i, t}^P < 0 \\[2mm]
        Y^*_{i,t} & \!\text{ otherwise}
    \end{array}\right. ,
\label{eq:adg-price-setting}
\end{align}
where $\rho$ is the quantity adjustment parameter and $\eta \sim U(0, \bar{\eta})$ is a uniform random variable whose range is fixed by the price adjustment parameter $\bar{\eta}$.

Basically, C-firms of the basis model only adjust their selling price towards the average of the market, and production to compensate for current demand.
In this sense, these firms can be considered bounded rational agents that act as trend followers.

\subsection{The RL agents}

We expand the basis model by introducing fully rational C-firms as \emph{RL agents} that learn their price-quantity strategies by interacting with the environment and maximising a reward function.
In particular, the underlying decision model for a subset $N$ of fully rational C-firms corresponds to a stochastic game (see e.g.\ \cite{ShapSG, vdWalSDP}) with $N$ players, i.e.\ the RL agents, or, in the special case when $N = 1$, a Markov decision process. 
The remaining bounded rational agents, those of the basis model, are not players of the game as they all act following the same fixed behavioural rules, and they can hence be seen as part of the probabilistic transition of the stochastic game.

Each RL agent aims to maximise the sum of its discounted rewards
\begin{equation}
    S_i = \sum_{t=1}^{t_\textit{sim}} \gamma^t r_{i,t}.
\end{equation}
The maximisation of rewards (to be later defined as firms' profits) is the feature of the RL agents by which we refer to them as \emph{fully rational}, in contrast with \emph{bounded rational} firms that follow a behavioural heuristic that is not the result of an explicit optimisation. 

We train each RL agent using a standard asynchronous \emph{Q-learning} algorithm, like that introduced in~\cite{WDQL}, with $\epsilon$ greedy policy.
Therefore, each RL agent considers discrete state and action spaces, namely $\mathcal{S}$ and $\mathcal{A}$, and learns a value function $Q : \mathcal{S} \times \mathcal{A} \rightarrow \mathbb{R}$, through the well-known Bellman equation, to estimate the expected reward $r$ for each state-action pair $(s, a)$.
Note that, as explained below, the agents do not have full knowledge of the state of the model, nor can they fully observe the actions performed by the other agents. 
Because of this, we resort to such a technique, where each agent regards the others as part of the environment.

\vspace{0.1cm}
\textbf{State space.}
Similarly to standard (bounded rational) firms, RL agents use the price delta $\tilde{\Delta}^P_{i, t}$ and the firm stock $\tilde{\Delta}^Y_{i, t}$ values as state observations $\mathcal{S}$. This means that the current state of the underlying stochastic game model is only partially observable by agents.
In the case of RL agents, the two quantities are defined by first taking the logarithms of the corresponding quantities
\begin{align} \tilde{\Delta}^P_{i, t} &= \log(P_{i, t}) - \log(P_t) = \log(P_{i, t} / P_t) \\
\tilde{\Delta}^Y_{i, t} &= \log(Y_{i, t}) - \log(Y_{i, t}^d) = \log(Y_{i, t}/Y_{i, t}^d).
\label{eq:log_price_delta}
\end{align}
Since tabular Q-learning requires a finite observation space, these quantities are binned in $n_{\mathcal{S}}$ equally spaced poles from a minimum $\mathcal{S}_{min}$ to a maximum $\mathcal{S}_{max}$.
Each real-valued observation is then assigned the index of the nearest pole. 
The logarithmic transformation in the state space definition above allows us to interpret the two states as the percentage differences of the past choices with respect to past market signals.

\vspace{0.1cm}
\textbf{Action space.}
Based on the current state observations, each RL agent $i$ can make price-quantity decisions for the next step, thus setting $P_{i, t+1}$ and $Y^*_{i, t+1}$, respectively. 
To change these values the agent selects an action set $a_{i, t} = (a_{i, t}^Y, a_{i, t}^P)$.
Given the selected actions, prices and quantities are adjusted as
\begin{equation} 
P_{i, t+1} = e^{\log(P_{i, t}) + a_{i, t}^P},  \quad \quad \qquad
Y^*_{i, t+1} = e^{\log(Y_{i, t}^*) + a_{i, t}^Y}.
\end{equation}
%
%
Similarly to the state observations, the actions are required to be discrete. 
We select a number $n_\mathcal{A}$ of discrete actions to be uniformly spaced between a minimum and maximum value, $\mathcal{A}_{min}$ and $\mathcal{A}_{max}$ respectively.

\vspace{0.1cm}
\textbf{Reward.}
At each step $t$, each RL agent observes $s_{i, t} = \left( \tilde{\Delta}^P_{i, t}, \tilde{\Delta}^Y_{i, t}, \right)$ performs action $a_{i, t} = \left( a_{i, t}^P, a_{i, t}^Y \right)$, and obtains a reward $r_{i,t}$ given by
\begin{equation}
    r_{i, t} = \left\{\!\!
    \begin{array}{ll}
        \pi_{i, t} & \text{ if } A_{i, t} > 0 \\[2mm]
        -100  & \text{ otherwise}
    \end{array}\right. ,
\end{equation}
where $\pi_{i, t}$ are the firm's profits and $A_{i, t}$ are the firm's assets.
Profits are computed as total revenues i.e., the quantity of sold goods $Y^s_{i, t}$ times the retail price $P_{i t}$, minus total costs, consisting of labour, investments, debt instalments and dividends.
All quantities are adjusted for inflation and hence expressed in real terms.
Basically, the rewards of the RL agent are the profits unless the firm goes bankrupt, where it gets penalised more heavily. 
Notice that, as we consider a multi-agent setting, each agent's reward actually depends on the current state and the joint actions of all the agents.

From a game-theoretical standpoint, the underlying stochastic game is guaranteed to admit optimal policies for the RL agents, given some solution concept such as Markov perfect equilibrium~\cite{MTMPEOA}. However, the corresponding problem is known to be at least PPAD-hard, since finding a Nash equilibrium is already PPAD-complete (see e.g.~\cite{DGP2009ComplexityNE}).
Therefore, while this could theoretically be done as the basis model can provide a full definition of the states, actions, transition and rewards, it would not be feasible in practice as the size of the model and of the strategy space are highly prohibitive.
Moreover, the possible use of information extracted from the game model would not represent a realistic assumption since, in the real world, agents are not able to fully observe their environment.
On the other hand, the adopted Q-learning procedure allows each agent to learn a policy which approximates the best response for the behaviour of bounded rational agents and, possibly, the other RL agents as well.

\vspace{0.1cm}
\textbf{Training.} 
During the training each agent $i$ updates the Q-function according to the following Bellman equation 
\begin{equation}
\resizebox{.91\linewidth}{!}{$%
    Q(s_{i,t}, a_{i,t}) \leftarrow (1 - \alpha) Q(s_{i,t}, a_{i,t}) + \alpha \left(r_{i,t} + \gamma \max_{a} Q(s_{i,t+1}, a)\!\right),%
$}
\label{eq-qlearning}
\end{equation}
where $\alpha$ is the learning rate, $\gamma$ is the discount factor and $s_{i, t+1}$ is the next state observed by the RL agent.
Considering the multi-agent setting, we differentiate between training with \textbf{shared policies} and with \textbf{independent policies}. 
In the former case, the agents share and update the same Q-matrix. 
In the latter case, each agent $i$ will use and update its own matrix $Q_i$.

During training, each agent follows an \emph{$\epsilon$-greedy} action selection
\begin{equation}
    a_{i, t} = \left\{\!\!
    \begin{array}{ll}
        \text{random action} & \text{ with probability } \epsilon \\[1mm]
        \arg\max\limits_{a} Q(s_{i,t}, a) & \text{ otherwise}
    \end{array}\right. ,
\label{eq:eps-greedy}
\end{equation}
with $\epsilon$ slowly decaying for progressing training episodes. 
Once trained, each agent follows the learned policy and chooses the action for which Q-matrix is maximised i.e., $\arg\max_{a \in \mathcal{A}} Q(s,a)$.

\begin{table}[t!]
    \centering
    \small
    \begin{tabular}{lll}
        \toprule
        \textbf{Symbol} & \textbf{Description} & \textbf{Value} \\
        \midrule
        $T_\textit{train}$ & Number of training episodes & $100$ \\
        $T_\textit{test}$ & Number of test episodes & $20$ \\
        $t_\textit{sim}$ & Number of steps for each simulation & $5000$ \\
        $t_\textit{burn-in}$ & Number of burn-in steps & $300$ \\
        $H$ & Number of workers & $1000$ \\
        $F_c$ & Number of C-firms & $100$ \\
        $F_k$ & Number of K-firms & $20$ \\
        $z_c$ & Number of C-firms visited by a consumer & $\{ 2, \dots, 10\}$ \\
        $N$ & Number of RL agents & $\{1, \dots, 20 \}$ \\
        $\gamma$ & Discount factor & $0.95$ \\
        $\alpha$ & Learning rate & $0.1$ \\
        \bottomrule
    \end{tabular}
\caption{\textbf{Experimental parameters used}.
\normalfont
List of parameters used in the simulations.
The specific values used for $z_c$ and $N$ can be read  directly in the figures of Section~\ref{sec:results}.
}
\label{tab:params}
\vspace{-0.2in}
\end{table}

\begin{figure*}[ht]
  \centering
    \includegraphics[width=1.0\linewidth]
    {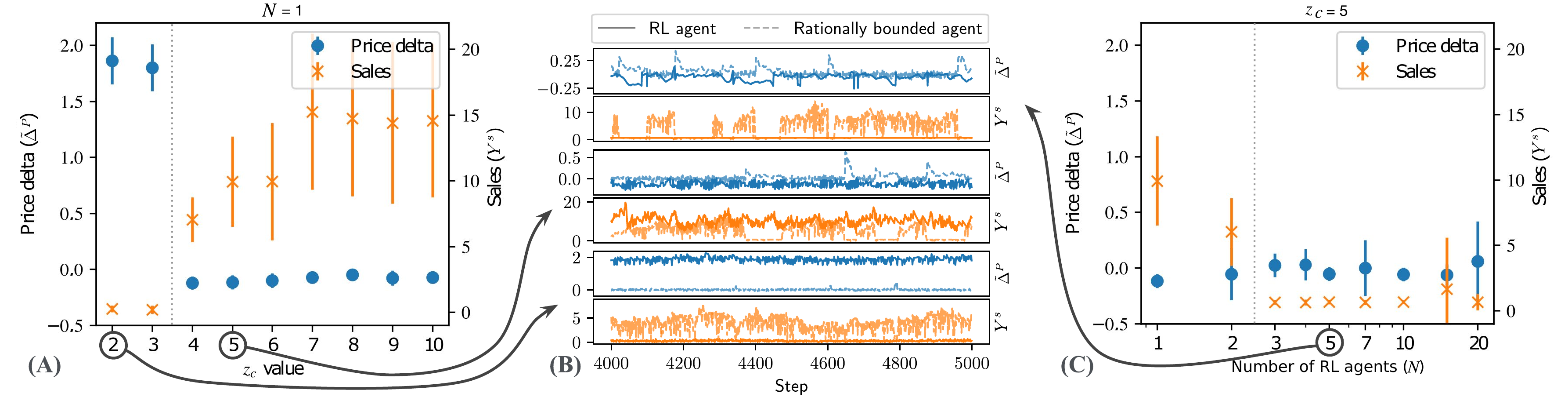}
       \vspace{-0.15in}
  \caption{\textbf{Different emerging strategies for different economic environments.}
  \normalfont
  The middle panel (B) shows the time series of observed sales ($Y^s$) and price-deltas ($\tilde{\Delta}^P$) for a sample bounded rational agent (dashed line) and RL agent (solid line) under 3 different combinations of market competition ($z_c$) and degree of rationality ($N$) as indicated by the arrows.
  The 3 combinations give rise to 3 different strategies for the RL agents. 
  From top to bottom, we find `\textbf{\emph{perfect competition}}', `\textbf{\emph{dumping}}' and `\textbf{\emph{market power}}' strategies (see main text for more details).
  The change in the emerging strategy for different economic environments is highlighted in the left (A) and right (C) panels, which plot the average value of price-delta (left axes) and sales (right axes) as a function of $z_c$ and $N$ respectively.
  }
  \label{fig:sales_price_zc}
\vspace{-0.1in}
\end{figure*}

\vspace{-0.1cm}
\section{Experimental setup} 
We simulate the macroeconomic model considering a single bank, $1000$ workers, $20$ K-firms, and $100$ C-firms. 
In our experiments, we vary the levels of competition ($z_c$) from $2$ to $10$, and the number of RL agents ($N$), representing fully rational C-firms, from $1$ to $20$. 
Since we focus exclusively on C-firms in our experiment, we will refer to them simply as `firms' for the remainder of the work.
All simulation parameters are described in Table~\ref{tab:params}, while additional parameters of the basis model are those of Assenza et al.~\cite{assenza2015emergent}.

We train the RL agents for each model configuration, considering both shared and independent policies. 
We consider $T_\textit{train}=100$ episodes for the training phase, a number which we have found more than sufficient for reaching a full convergence of the Q-learners in all our experiments, and consider $T_\textit{test}=20$ episodes for testing. 
Unless otherwise specified, we report averages of the simulation results over the 20 testing episodes, with error bars representing one standard deviation. 

Similarly to some existing MARL frameworks for economic modelling, we consider a sort of \textit{curriculum learning}~\cite{bengio2009curriculum}: the simulation starts without the RL agents, which are introduced after $t_\textit{burn-in}$ steps.
Then, the RL agents assume control of $N$ firms for additional $t_\textit{sim}$ steps.
At the start of the training phase, RL agents are initialised with an empty Q-matrix i.e. all values are zero. 
During training, and  at each episode $\tau = 1, \dots, T_\textit{train}$ , the agents update the exploration rate $\epsilon$ of the  $\epsilon$-greedy policy in Eq.~(\ref{eq:eps-greedy}) as
 \mbox{$\epsilon \leftarrow \max \left( 0.9^{\tau-1}, 0.01 \right)$}.

\vspace{-0.1cm}
\section{Results}\label{sec:results}

We now discuss the simulations obtained with our R-MABM framework, with standard firms gradually substituted by RL agents trained to maximise profits. 
We divide our results into the microeconomic effects (Section \ref{subsec:results_micro}) and the macroeconomic effects (Section \ref{subsec:results_macro}) of the introduction of rational RL agents.

\subsection{The microeconomic impact of RL agents}
\label{subsec:results_micro}
We first evaluate the microeconomic impact of RL agents learning to efficiently produce and price goods in competition with each other and with bounded rational firms.
For each firm, we measure its \textit{sales} $Y^s$ of consumption goods and its \emph{price delta} $\tilde{\Delta}^P$, to assess the distance between the firm's price and the average market price. 

In general, we find that RL agents adapt their strategy to the level of market competition $z_c$ and rationality $N$, outperforming the profits of bounded rational firms from the basis model.

\vspace{0.1cm}
\textbf{Different emerging strategies for different levels of competition.}
We first evaluate the impact of different levels of competition through the model parameter $z_c$, considering a simulation with a single RL agent ($N=1$). 
Figure~\ref{fig:sales_price_zc}.A shows the agent policy in terms of price delta and sales. 
We observe that at increasing levels of market competition, the RL agent exhibits different strategies. 

When the competition is low, for $z_c$ 2 or 3, the RL agent learns to maximise profits by producing very low quantities of goods and selling them at skyrocketing prices. 
We call this strategy a \textit{``market power''} strategy since the RL agent has learned that he can charge any desired price on the goods sold and always find buyers since consumers are forced to visit only a small number of firms to cover their consumption needs.

When the competition in the market gets higher, for $z_c\geq 3$, the market power strategy becomes less profitable, and the learned price-quantity strategy changes abruptly.
In these circumstances, the RL agent learns to drop the retail price below market level in order to undercut the competition, and it further learns that by charging lower prices it can produce and sell higher quantities.
This, in turn, allows the RL agent to secure a large portion of the market share and accumulate high profits.
We call this strategy a \textit{``dumping''} strategy.

\begin{figure*}[t!]
  \centering
  \includegraphics[width=1.0\textwidth]{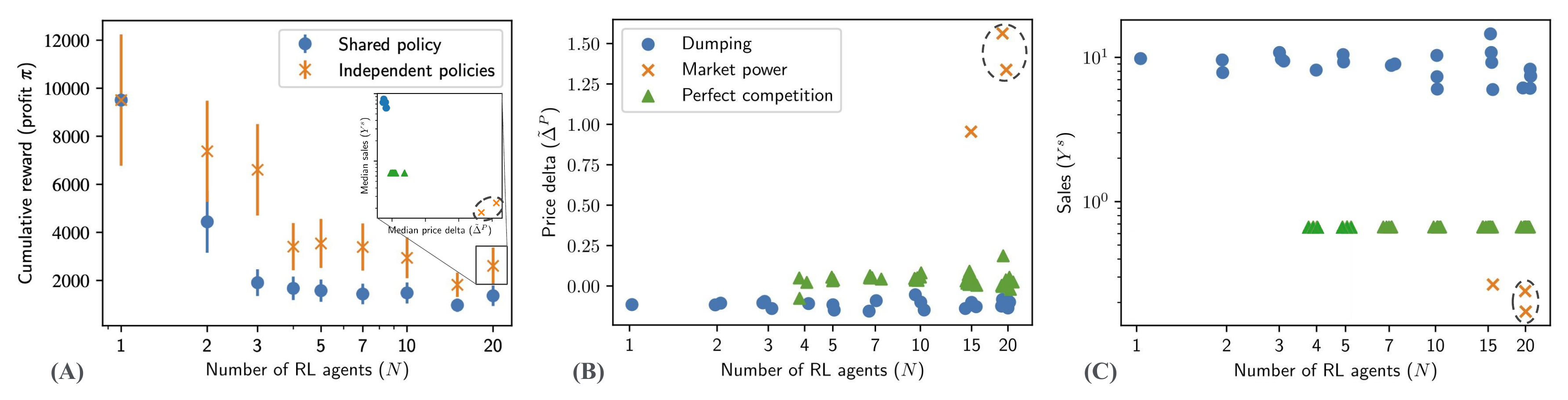}
    \vspace{-0.3in}
  \caption{\textbf{Independent RL-agents spontaneously segregate into strategic groups increasing overall profits.}
  \normalfont
  The left panel (A) shows the total mean cumulative rewards (i.e., profits) for a varying number of agents, with $z_c=5$, and for RL agents with shared or independent policies.
  Agents with independent policies always achieve higher overall rewards as a result of spontaneous segregation into strategic groups.
  Segregation can be clearly observed in the middle and right panels (B) and (C), showing the value of agent-specific price delta and sales as a function of a growing number of rational agents, with $z_c=5$. 
  A small noise was added to the $x$-axis to better resolve very nearby points.
  The different strategic groups are particularly easy to spot when plotting price delta against sales, as done in the inset of (A) for $N=20$.
  }
  \label{fig:compared_rewards}
  \vspace{-0.1in}
\end{figure*}

Figure~\ref{fig:sales_price_zc}.B shows some insights from simulations with $z_c =2$ and $z_c=5$, comparing the RL agent strategies against a bounded rational firm.
For the \textit{``market power''} strategy ($z_c =2$), we clearly see that the RL agent charges much higher retail prices, and this implies much lower sales than the bounded rational firm.
In spite of the very low sales, the strategy gives rise to high total revenues ($P_t \cdot Y^s_t$) and concomitantly to low total costs, yielding high profits.

The figure shows an even more interesting behaviour for the RL agent with the \textit{``dumping''} strategy ($z_c = 5$). 
In fact, the RL agent adaptively changes the prices of goods in a very sophisticated manner to keep them just slightly below the competitors' prices. 
This is evident from the fact that the solid blue line, representing the RL agent price, is always below the dotted light blue line, representing the price charged by a sample bounded rational firm.
The low prices allow for high sales, as demonstrated by the solid orange line being always considerably higher than the dashed orange line.

For both strategies, the RL agent has also successfully learned to avoid bankruptcy.
This can be best observed from the time series of sales for the dumping strategy, as the sales of the RL agents remain high for the entire time while the sales of the bounded rational firm sometimes drop to zero for a certain number of steps, indicating a sudden bankruptcy and a subsequent slow recover.

\vspace{0.1cm}
\textbf{Emerging perfect competition for multiple fully rational agents with a shared policy.} 
In Figure~\ref{fig:sales_price_zc}.C we investigate the case of multiple RL agents, with competition level fixed at $z_c=5$. 
In this setting, we train $N$ RL agents using a \textit{shared} policy, i.e., all agents share and update the same Q-matrix.

While for $N=1,2$ the RL agents learn the dumping strategy discussed in the previous paragraph, a new transition appears after $N=2$ and the strategy learned by three or more RL agents is entirely new.
In this scenario, the even higher level of competition caused by the presence of multiple fully rational firms in the economic system drives the RL agents to learn the very conservative strategy of producing moderate amounts and selling them at a price which is in line with the market.
By stretching the technical meaning of the term, we call this strategy `perfect competition', since under this strategy RL firms have no power to affect prices. 
They are `price takers' and closely follow consumer demand.
Notably, perfect competition is \emph{not} optimal for low levels of competition ($z_c$) and for only a few fully rational firms ($N$), when instead dumping or market power strategies are much more profitable; but it emerges as the only viable strategy when the economic system is composed of many fully rational firms competing fiercely against each other for sustained profits and avoidance of bankruptcy.

In Figure~\ref{fig:sales_price_zc}.B we show an insight into the simulation with $N=5$, showing the strategy of a sample RL agent, which typically sells fewer goods than a bounded rational agent, at similar or lower prices.
It is again interesting to notice the adaptive pricing behaviour of the RL agent from the top inset of Figure~\ref{fig:sales_price_zc}.B.

Finally, we note that average profits of RL agents diminish substantially in the transition from an environment with $N=1,2$, which can be exploited with a dumping strategy, to an environment with $N\geq 3$, where RL agents are forced into perfect competition to stay profitable and bankruptcy-free. 
This abrupt change in profits is reported in Figure~\ref{fig:compared_rewards}.A.
The figure also illustrates that profits keep decreasing although more gradually also for $N\geq 3$, in line with classical microeconomic theory.
Although we do not explicitly show the profits/rewards of bounded rational firms, we note here that these are consistently lower than those of RL agents as illustrated for a single run in the right panel of Figure~\ref{fig:training_reward}.

In summary, RL agents spontaneously find the following three strategies to maximise profits:
\vspace{-0.1cm}
\begin{itemize}
    \item \textbf{Market power.} Firms maximise profits by exploiting the imperfection of the market, with households being able to visit only a small number of firms for their consumption. 
    They produce and sell small amounts of goods charging very high prices, thus at the same time maximising revenues and minimising costs.
    \item \textbf{Dumping.} Firms maximise profits by dropping their prices below market level to eliminate competition and at the same time producing as much as possible to increase sales and gain a significant market share.
    \item \textbf{Perfect competition.} Firms achieve sustainable profits and avoid bankruptcy by a conservative strategy that involves producing and selling moderate amounts of goods, and charging market level prices.
\end{itemize}

\begin{figure*}[t]
    \centering
    \includegraphics[width=0.99\textwidth]{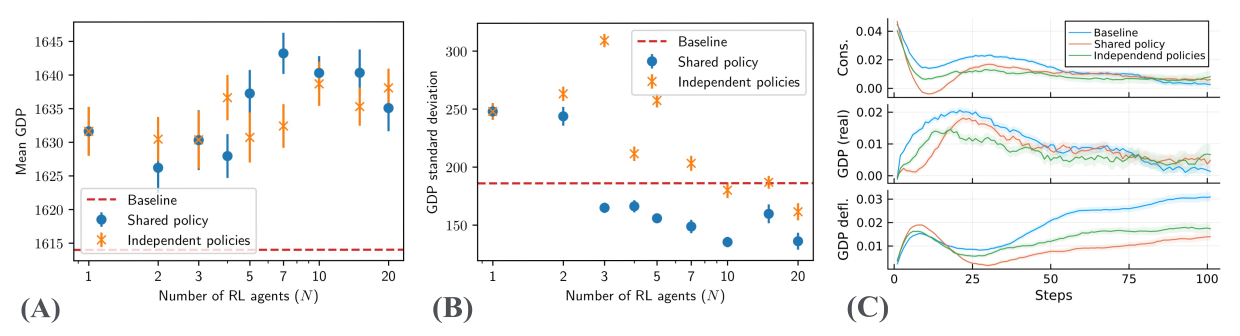}
        \vspace{-0.15in}
    \caption{%
    \textbf{
    RL agents always increase total macroeconomic output, but only perfect competition also increases economic stability.} 
    \normalfont
    Mean (A), and standard deviation (B) of the GDP in the last 1000 steps of simulations with a growing number of rational agents, trained either with shared or independent policies.
    Error bars represent the standard error on average quantities. 
    The last panel (C) shows the response function of consumption, GDP and GDP deflator to a positive shock in the propensity to consume of all households.
    All results of the figure are obtained with $z_c=5$ and the impulse responses are obtained with $N=7$.
    }
    \label{fig:gdp}
    \vspace{-0.1in}
\end{figure*}

\vspace{0.1cm}
\textbf{Independent RL agents increase profits over perfect competition by segregation.}
We now investigate the case of multiple RL agents trained with \textit{independent} Q-matrices, and hence free to learn different policies.
Figure~\ref{fig:compared_rewards}.A shows the cumulative average rewards achieved by an increasing number $N$ of RL agents at fixed $z_c=5$, for training with both \textit{independent} and \textit{shared} policies. 
The first immediately visible emergent phenomenon is the fact that RL agents with individual policies achieve higher total profits/rewards on average as compared with RL agents that are constrained to the same common policy.
While this may seem counterintuitive, since an agent would maximise its individual rewards at the expense of the others, we recall that our model is not a zero-sum game.
Independent agents can achieve higher profits by adjusting their strategies and holding different market niches.
Specifically, we here observe an interesting phenomenon of spontaneous segregation of the RL agents into different strategic groups, where agents within each group follow one of the three strategies described above.
This, in turn, increases profits substantially over the shared policy scenario since certain strategic groups can still carry on exploitative behaviour which would not be sustainable for all RL agents at the same time.
The spontaneous segregation of agents is particularly clear from the inset of Figure~\ref{fig:compared_rewards}.A, where the 20 RL agents are seen to cluster into all three discovered strategies according to their median sales and price.

Panels B and C of Figure~\ref{fig:compared_rewards} show the gradual emergence of the segregation into strategic groups, which begins at $N=4$ with a separation between dumping and perfect competition, and continues at $N=15$ with the addition of the market power group.
Interestingly, while RL agents with a shared policy transition to perfect competition already for $N>2$, RL agents with independent policies are able to continue leveraging the dumping policy even with $N=3$, as a result of subtle tacit coordination in the three dumping strategies enacted.
This results in drastically higher profits as clearly observable from Figure~\ref{fig:compared_rewards}.A.

\subsection{The macroeconomic impact of RL agents}
\label{subsec:results_macro}
In this section, we investigate the macroeconomic effects of introducing RL agents into our R-MABM.
In general, we find that a higher degree of rationality in the economy always improves the macroeconomic environment as measured by total output, but that the macroeconomic stability of the system depends on the specific strategy put in place by the rational firms.

\vspace{0.1cm}
\textbf{Increased rationality implies higher overall output.}
Figure~\ref{fig:gdp}.A shows the average gross domestic product (GDP) of simulations with RL agents with shared or independent policies for increasing~$N$.
The figure clearly shows that, independently of the type of policy, any number of RL agents $N$ gives rise to a significant increase in the overall output of the economic system with respect to the basis model. 
This is not surprising since rational firms have a greater incentive to increase production levels to increase profits.
This happens both in `exploitable' economic environments that allow few RL agents to gain a high market share via dumping ($N=1,2$) and in `unexploitable' markets where RL agents are forced into perfect competition and achieve higher outputs collectively.

\vspace{0.1cm}
\textbf{Increased rationality implies higher economic stability only when it also implies perfect competition.}
Figure~\ref{fig:gdp}.B shows the economic instability of the R-MABM, measured by the standard deviation of the GDP, for an increasing number $N$ of RL agents trained with shared or independent policies.
While overall output always increases with increasing rationality, economic stability does not necessarily increase.
Specifically, we find that rationality increases economic stability only when it also implies the adoption of the perfect competition strategy by RL firms.
On the contrary, RL firms adopting an exploitative behaviour like the dumping strategy, capitalise on firms with limited rationality making them more susceptible to bankruptcy, thus increasing overall economic instability.

The effect can be clearly seen by noticing how the instability of share policy RL agents is higher than in the basis model for $N=1,2$, when damping is the optimal strategy, and how it drops rapidly to below baseline for $N\geq3$, when RL agents are forced into perfect competition.
Moreover, independent policy RL agents keep exploiting the dumping strategy as described in the previous section, and hence consistently give rise to higher economic instability than shared policy RL agents.
Also in the case of independent policies the instability is seen to decrease for larger values of $N$, but in a more gradual fashion, as a result of perfect competition gradually becoming the prevalent strategy.

\vspace{0.1cm}
\textbf{Perfect competition gives rise to the highest overall output and to efficient responses to aggregate shocks.}
The highest overall outputs are achieved at $N=7,10,15$ for RL agents with shared policy following a perfect competition strategy, in spite of the fact that the dumping strategy implies higher production levels for rational firms.
The perfect competition strategy also appears to give rise to quick and effective responses to aggregate shocks, as exemplified in Figure~\ref{fig:gdp}.C.
The figure shows the impulse response functions for three macroeconomic variables: consumption, real GDP, and GDP deflator (a measure of inflation).
The impulse (or `shock') consists of an instantaneous 30\% increase in the propensity to consume of all households and is computed for the baseline model and for R-MABM models with shared/independent policies.
The curves are obtained following the technique described in \cite{gatti2020rising}.

Notably, the rational models are able to more quickly satisfy the increased demand as indicated by the more rapid convergence of the shocked consumption variable to the pre-shock value (zero in the graphs). 
At the same time, they also generate lower inflation as measured by the GDP deflator variables.
The efficiency in the economic response is particularly strong for the shared policy model and could be a direct consequence of the perfect competition strategy which is fully adopted only by shared policy RL~agents.

\vspace{-0.2cm}
\section{Conclusions}

In this work, we introduce the use of multi-agent reinforcement learning to extend traditional macroeconomic agent-based models (ABMs).
Specifically, we propose a `Rational macro ABM’ (R-MABM) framework that allows substituting firms in a classical macro ABM with profit-maximising reinforcement learning (RL) agents.
We define RL agents to be `fully rational' as they explicitly optimise a reward function, and contrast them with `bounded rational' agents with fixed behavioural rules, hence the name of the framework. 
We showcase how our framework allows for modelling and studying increased levels of rationality (the number of RL agents) in the economy, a problem that cannot be investigated with traditional macro ABMs.

We find a number of empirical results that can be well connected with ideas and concepts from economic theory.
On the microeconomic level, we find that RL agents spontaneously learn three distinct price-quantity strategies to maximise profits depending on the economic environment they are immersed in.
When competition is very low, they learn a `market power’ strategy consisting of charging very high prices and producing very little, relying on the fact that with low competition there is always demand for their goods.
When competition is higher but there are few RL firms, they learn a `dumping’ strategy consisting of flooding the market with high quantities of goods sold at very low prices, thus undercutting competition from the bounded rational agents.
Finally, when both competition and the number of RL firms in the economy are high, RL agents are forced into a `perfect competition’ strategy consisting of producing enough goods to satisfy the demand and selling them at market prices.

Notably, we also find that when RL agents are allowed to follow independent policies they spontaneously learn to segregate into different strategic groups thus increasing profits for all.

On the macroeconomic level, we find that RL firms always improve overall economic output, but that macroeconomic stability is improved only when they are forced into a strategy of perfect competition.
We further find that RL agents in perfect competition strategy typically give rise to the greatest economic output and to high economic stability and resilience to shocks.

This work is a first step towards the investigation of the use of RL in traditional economic ABMs and, more generally, an important step towards the integration of artificial intelligence in economic simulation models.
Our R-MABM framework allows for stable multi-agent learning, and thus it provides a solid foundation for many possible extensions. 
Future work could involve increasing the action space of rational firms to include other choices such as investments, or studying emergent decision-making from RL models of other agent classes in the basis model, such as the government, the central bank, the commercial banks or the households.

\begin{acks}
We would like to thank Adrián Carro (Bank of Spain), Herbert Dawid (Bielefeld University) and Marco Pangallo (CENTAI institute) for constructive feedback on this work.

The views and opinions expressed in this paper are those of the
authors and do not necessarily reflect the official policy or position
of Banca d’Italia.
\end{acks}

\section*{Code availability}
In the interest of reproducibility, the code to simulate the R-MABM
framework is available in open source at \url{https://github.com/Brusa99/R-MABM}. 
The code is based on the MABM model as available at \url{https://github.com/bancaditalia/ABCredit.jl}.

\balance

\bibliographystyle{ACM-Reference-Format.bst}

\bibliography{acm-ec}

\end{document}